\documentclass[runningheads]{paper654}

\usepackage{graphicx}
\usepackage{multirow}
\usepackage{booktabs}
\usepackage{amsmath}
\usepackage{amssymb}

\usepackage{bbding}
\usepackage[pagebackref=true,breaklinks=true,colorlinks,bookmarks=false]{hyperref}

\begin{document}
\title{ORF-Net: Deep Omni-supervised Rib Fracture Detection from Chest CT Scans}
\titlerunning{ORF-Net: Deep Omni-supervised Rib Fracture Detection}
%

\authorrunning{Zhizhong Chai, et al.}

\author{Zhizhong Chai\inst{1}\Envelope,
Huangjing Lin\inst{1},
Luyang Luo\inst{2},\\
Pheng-Ann Heng\inst{2,3},
\and Hao Chen\inst{4}}

\institute{$^1$Imsight AI Research Lab, Shenzhen, China\\
\email{chaizhizhong@imsightmed.com}\\
$^2$Department of Computer Science and Engineering,\\
The Chinese University of Hong Kong, Hong Kong, China\\
$^3$Guangdong Provincial Key Laboratory of Computer Vision and Virtual Reality Technology, 
Shenzhen Institutes of Advanced Technology, Chinese Academy of Sciences, Shenzhen, China\\
$^4$Department of Computer Science and Engineering,\\
The Hong Kong University of Science and Technology, Hong Kong, China
}

\maketitle 
\begin{abstract}
Most of the existing object detection works are based on the bounding box annotation: each object has a precise annotated box. However, for rib fractures, the bounding box annotation is very labor-intensive and time-consuming because radiologists need to investigate and annotate the rib fractures on a slice-by-slice basis. Although a few studies have proposed weakly-supervised methods or semi-supervised methods, they could not handle different forms of supervision simultaneously. In this paper, we proposed a novel omni-supervised object detection network, which can exploit multiple different forms of annotated data to further improve the detection performance. Specifically, the proposed network contains an omni-supervised detection head, in which each form of annotation data corresponds to a unique classification branch. Furthermore, we proposed a dynamic label assignment strategy for different annotated forms of data to facilitate better learning for each branch. Moreover, we also design a confidence-aware classification loss to emphasize the samples with high confidence and further improve the model's performance. Extensive experiments conducted on the testing dataset show our proposed method outperforms other state-of-the-art approaches consistently, demonstrating the efficacy of deep omni-supervised learning on improving rib fracture detection performance.

\end{abstract}

\section{Introduction}
Rib fractures are the most common disease in thoracic injuries \cite{talbot2017traumatic}. Although most fractures only require conservative treatment, the complications caused by rib fractures can be excruciating to patients \cite{sirmali2003comprehensive}. 
In addition, the mortality rate will rise with the increasing number of rib fractures \cite{talbot2017traumatic}.
Therefore, the accurate recognition and location of rib fracture have important clinical value for patients with thoracic trauma.
However, the identification of rib fractures in Computed Tomography (CT) images is tedious and labor-intensive, especially for subtle rib fractures and buckle rib fractures, resulting in missed and misdiagnosed rib fractures in clinical practice \cite{cho2012missed,ringl2015ribs}.
Recently, deep learning (DL) has achieved comparable performance with the experienced radiologists \cite{weikert2020assessment,wu2021development,zhou2020automatic}, based on enormous annotated data.
Although detailed annotations are essential to DL-based disease detection \cite{luo2021rethinking}, they are time-consuming and expertise-demanding, which is a considerable challenge for annotating rib fractures.

To ease the dependence on expensive annotated data, some studies have proposed methods \cite{jeong2019consistency,liu2021unbiased,sohn2020simple,wang2020focalmix} based on weakly supervised learning or semi-supervised learning.
For example, Wang et al. \cite{wang2021knowledge} proposed an adaptive asymmetric label sharping algorithm to address the class imbalance problem for weakly supervised object detection. 
Zhou et al. \cite{zhou2021ssmd} developed an adaptive consistency cost function to regularize the predictions from different components.
However, practical applications are usually faced with a variety of different supervision labels.
To address the above issue, omni-supervised learning \cite{radosavovic2018data} was proposed to leverage different types of available labels to jointly train the model and boost the performance. 
For example, Ren et al. \cite{ren2020ufo} proposed a unified object detection framework that can handle several different forms of supervision simultaneously. Luo et al. \cite{luo2021oxnet} presented a unified framework that can simultaneously utilize strong-annotated, weakly-annotated, and unlabeled data for chest X-ray disease detection. However, both of the above methods are based on the anchor-based detection networks, which have many hyper-parameters that require careful design and tuning.

A potential challenge of omni-supervised learning is that it requires designing reliable label assignment strategies for different forms of annotated data, so as to enable the model to be effectively trained with different supervision.
The label assigning strategies commonly used in exiting object detection works can be separated into two categories: fixed label assignment and dynamic label assignment. The fixed label assignment-based methods \cite{ren2015faster,tian2020fcos} are heavily based on hand-crafted rules. For example, Faster-RCNN \cite{ren2015faster} uses the IoU thresholds to define the positive and negative proposals generated from the region proposal network (RPN). FCOS \cite{tian2020fcos} regards the pixels in the center region of the object bounding box as positives with a centerness score to refne the confdence. The dynamic label assignment-based methods \cite{2020Bridging,zhu2020autoassign} propose adaptive mechanisms to determine the positives and negatives. ATSS \cite{2020Bridging} adaptively selects positive and negative samples according to the statistical characteristics of objects. AutoAssign \cite{zhu2020autoassign} proposed an appearance-aware and fully differentiable weighting mechanism for label assignment in both spatial and scale dimensions. 

Inspired by the above dynamic label assignment-based methods for object detection, we design an omni-supervised framework that can utilize different annotation forms of data in a label assignment manner.
Specifically, the proposed omni-supervised network contains an omni-supervised detection head to handle the training of different annotated forms of data. 
Furthermore, to dynamically conduct the label assignment for each annotation form of data, we design a label assignment strategy that uses the prediction maps from the multiple classification branches as guidance. 
Moreover, a confidence-aware classification loss is also introduced to emphasize the samples with high confidence and suppress the noise candidates. 
Extensively experimental results show our proposed method outperforms other state-of-the-art approaches consistently, demonstrating the efficacy of deep omni-supervised learning on the task of rib fracture detection.

\section{Method}
\begin{figure}[t]
\centering
	\includegraphics[width=0.99\textwidth]{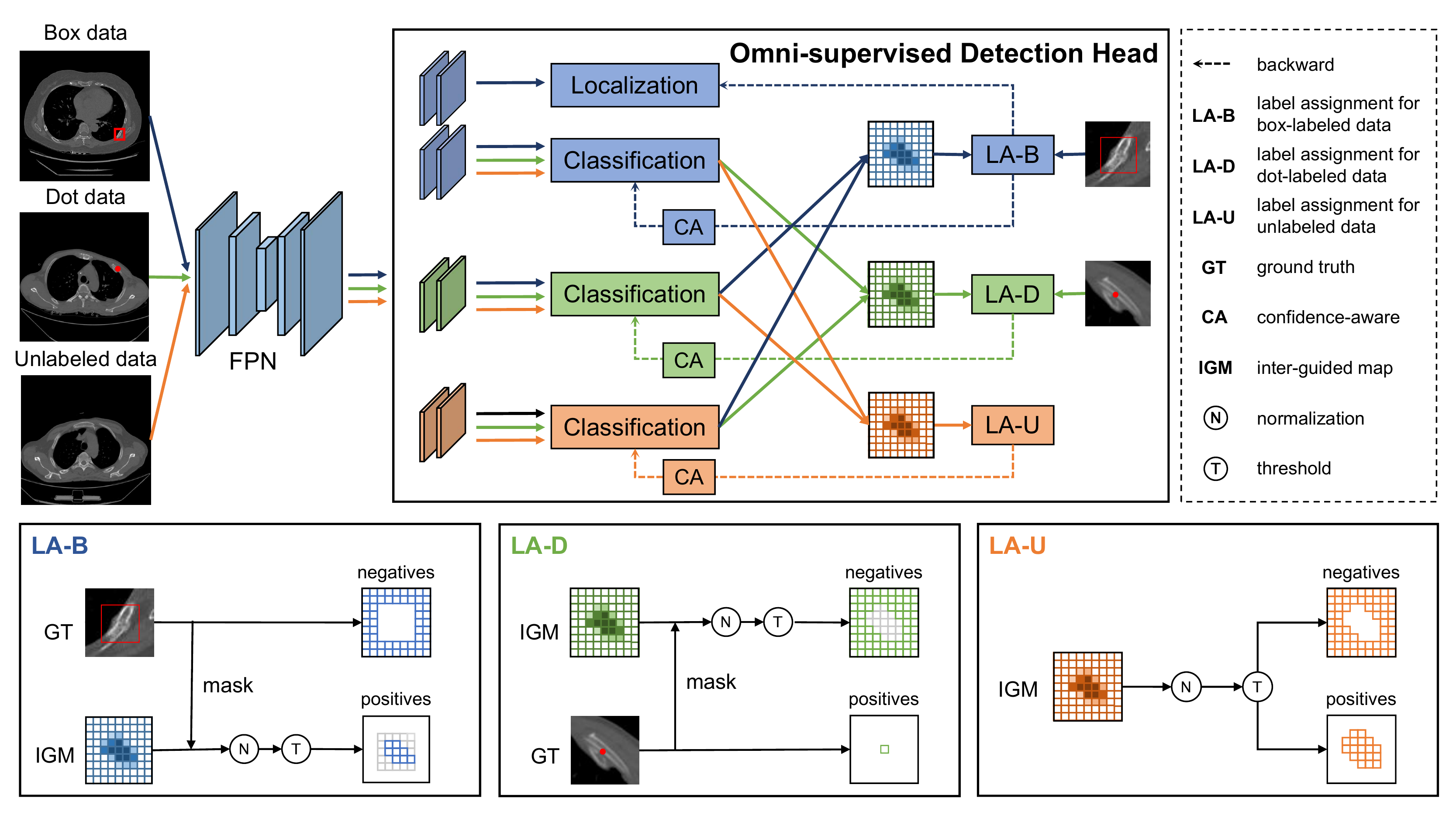}
	\caption{Overview of our proposed framework. The network consists of a Feature Pyramid Network (FPN \cite{lin2017feature}) as the backbone, and an omni-supervised detection head to predict the classification score and localization information. For each form of annotated data, there is a corresponding classification branch that is trained using a dynamic label assignment strategy. The proposed confidence-aware classification loss is adopted on the uncertain regions of different annotated forms of data.  
}
\label{framework}
	
\end{figure}
\subsection{Problem Statement and Formulation}
Formally, let us define the training dataset as ${D}$ = $[\mathcal{D}_{b}, \mathcal{D}_{d}, \mathcal{D}_{u}]$, where $\mathcal{D}_{b}$ is a bounding box annotated dataset, $\mathcal{D}_{d}$ is a dot annotated dataset with a single point for each object, and $\mathcal{D}_{u}$ is an unlabeled dataset. 
Intuitively, there are two types of the region for each form of labeled data: certain and uncertain.
For box-labeled data, the region outside the labeled boxes can be regarded as certain negative samples to supervise the model.
However, for the regions inside the labeled boxes, we cannot directly define their positive and negative attributes. 
If we simply treat all the samples in the box as positive samples, it will undoubtedly bring learning difficulties and performance degradation to the model \cite{zhu2020autoassign}.
Similarly, for dot-labeled data, the labeled points can be regarded as certain positive samples, but the locations in other regions have no explicit positive and negative attributes.
In this paper, we aim to design a label assignment strategy that can dynamically select high-quality samples from these uncertain regions during training for omni-supervised object detection.

Specifically, we build our model based on the one-stage detection network FCOS \cite{tian2020fcos}, which predicts the objects in an anchor-free manner. Fig. \ref{framework} illustrates the overview of our proposed method. 
The proposed omni-supervised detection framework consists of a Feature Pyramid Network (FPN \cite{lin2017feature}) and an omni-supervised detection head.
The omni-supervised detection head contains a localization head to regress the localization offsets, and multiple classification branches to predict the probability score for different annotated forms of data.
Since dot-labeled data and unlabeled data do not have specific labeling boxes, we only train the localization branch with the box-labeled data. 

\subsection{Omni-supervision with Dynamic Label Assignment}
To enable effectively learning from different labeled data and unlabeled data, we design a dynamic label assignment strategy that combines the prediction scores from the multiple classification branches and its own annotations to guide the model's training. 
Inspired by the success of co-training \cite{blum1998combining,xia2020uncertainty}, which minimizes the divergence by assigning pseudo labels between each view on unlabeled data, we use the prediction maps from the other branches to generate the confidence map, which we called the inter-guided map (IGM). By this mutual supervision mechanism, we could prompt different branches to maximize their agreement on different forms of annotated data for better performance. 
Suppose ${P}_{b}$, ${P}_{d}$, ${P}_{u}$ represent the predictions of the box-labeled branch, the dot-labeled branch, and the unlabeled branch, respectively, we calculate the inter-guided map with the prediction maps from the other two branches as follows: 
\begin{equation}
    {W}_{b} = N(({P}_{d}*{P}_{u})^\frac{1}{2}) \text{   , } {W}_{d} = N(({P}_{b}*{P}_{u})^\frac{1}{2}) \text{   , and }{W}_{u} = N(({P}_{b}*{P}_{d})^\frac{1}{2})\label{ia:inter-label assignment}
\end{equation}
where N denotes the normalization function which linearly rescales the values to the range of $[0, 1]$ and ensures effective learning of hard instances which usually have a small value for the corresponding positive locations. Note that we only use the uncertain region to generate the corresponding IGM map. For box-labeled data, the region in each annotated box is normalized separately. 
We then discuss the label assignment strategy for each annotated form of data separately.\\

\noindent\textbf{Box Supervision:}
Different from FCOS \cite{tian2020fcos} which uses a fixed fraction of the center area as positive samples for each object, we use a prediction-guided map to dynamically filter the locations of the bounding box as the reliable positive samples to train the model. 
Specifically, given an annotated bounding box $j$, we first calculate its iter-guided map $W_b^j$, and then we assign the location whose probability is greater or equal to the threshold $t$ in $W_b^j$ as a positive sample to train the box-labeled classification branch. For the negative regions $S_n$ outside the labeled boxes of the box-labeled data, the focal loss \cite{lin2017focal} is adopted to handle the severe imbalance problem. The positive loss and negative loss for box annotated data are computed as follows:
\begin{equation}
    \mathcal{L}_{b}^{p} = -\sum_{j}^{m}\sum_{W_b^{ij}>=t}^{S_p^j}(1-P_b^{ij})^{\gamma}\log{P_b^{ij}}
\end{equation}

\begin{equation}
    \mathcal{L}_{b}^{n} = -\sum_{i}^{S_n}(P_b^i)^{\gamma}\log{(1-P_b^i)}
\end{equation}
where $m$ denotes the number of annotated boxes in the slice, i denotes the i-th location from the FPN feature maps, and $S_p^j$ denotes the number of locations in the bounding box $j$.
For the localization branch, we use the generalized IoU (GIoU) \cite{rezatofighi2019generalized} loss (${L_{GIoU}}$) as the objective for bounding box localization:
\begin{equation}
    \mathcal{L}_{b}^{reg} = \sum_{j}^{m}\sum_{W_b^i>=t}^{S_p^j}{L_{\rm GIoU}(k_i,\hat{k}_i)}
\end{equation} 
where ${k}$ and $\hat{k}$ denote the predicted bounding boxes and the corresponding ground-truth boxes.\\

\noindent\textbf{Dot Supervision:}
For dot-annotated data, we assume the annotated locations are certain positive samples, and the other locations are uncertain regions where we should conduct the label assignment strategy. Specifically, we use the focal loss \cite{lin2017focal} to train the model on the annotated points. For the locations beyond the labeled points, We first calculate the inter-guided map according to Eq. (\ref{ia:inter-label assignment}) and then treat the locations whose probability is smaller than the threshold $t$ as negative samples to train the dot classification branch. 
The positive loss and negative loss for dot-labeled data are computed as follows:
\begin{equation}
    \mathcal{L}_{d}^{p} = -\sum_{i}^{l}(1-P_d^i)^{\gamma}\log{P_d^i}
\end{equation}

\begin{equation}
    \mathcal{L}_{d}^{n} = -\sum_{W_d^i<t}^{R_n}(P_d^i)^{\gamma}\log{(1-P_d^i)}
\end{equation}
where $l$ denotes the number of annotated dots in the slice, and $R_n$ denotes the number of unlabeled locations on the dot-labeled feature maps. \\

\noindent\textbf{Unlabeled Supervision:}
Since there is no certain region in the unlabeled data, we directly conduct the label assignment strategy on all locations on the feature maps. Specifically, we compute the inter-guided map of the locations and assign the locations whose probability is greater or equal to the threshold $t$ as positive samples, the others as negative samples. The loss function is as follows:
\begin{equation}
    \mathcal{L}_{u}^{p} = -\sum_{W_u^i>=t}^{T_p}(1-P_u^i)^{\gamma}\log{P_u^i}
\end{equation}
\begin{equation}
    \mathcal{L}_{u}^{n} = -\sum_{W_u^i<t}^{T_n}(P_u^i)^{\gamma}\log{(1-P_u^i)}
\end{equation}
where $T_p$ and $T_n$ denote the number of assigned positive samples and negative samples, respectively.  \\

\noindent\textbf{Confidence-aware Classification Loss:}
Although the above label assignment strategy can enable the model to be trained on data with different annotation types, directly dividing the positive and negative samples with a threshold may limit the detection performance of the model. To address this problem, we propose a confidence-aware (CA) classification loss for the uncertain regions of each annotated form of data to further emphasize the locations with high confidence. The classification losses are computed as follows: 

\begin{equation}
    \mathcal{L}^{p} = -\sum_{W^i>=t}^{S}(1-P^i)^{\gamma}\log{(P^i(1-W^i))}
\end{equation}
\begin{equation}
    \mathcal{L}^{n} = -\sum_{W^i<t}^{S}(P^i)^{\gamma}\log{(1-P^i(1-W^i))}
\end{equation}
where $W$ and $P$ denote the corresponding inter-guided map and prediction map, and $S$ denotes the number of locations on the FPN feature maps.

\subsection{Optimization and Implementation Details}
The final training loss is defined as follows:
  \begin{equation}
 \mathcal{L}_{total} = \mathcal{L}_{b}^{reg}+\mathcal{L}_{b}^{p}+\mathcal{L}_{b}^{n}+\lambda(\mathcal{L}_{d}^{p}+\mathcal{L}_{d}^{n})+\beta(\mathcal{L}_{u}^{p}+\mathcal{L}_{u}^{n})
 \end{equation}
where $\lambda$ and $\beta$ are empirically set to 1.
We use FCOS \cite{tian2020fcos} with ResNet-50 \cite{he2016deep} backbone pre-trained on ImageNet \cite{deng2009imagenet} and FPN as our base model. All experiments are conducted based on Pytorch \cite{paszke2019pytorch}. During training, we adopt 1 TITAN Xp GPU with a batch size of 3, where the three different annotation types of data are equally sampled. During the testing phase, we first ensemble the prediction results of different classification branches, and then combine the regression result from the localization branch to generate the final detection result. Random flip is applied for data augmentation. We use Stochastic Gradient Descent (SGD) with a momentum of 0.9 to update the weight of the model, and the initial learning rate is 0.001 and multiplied by 0.1 every 30000 iterations. We set threshold $t$ as 0.5 for all the annotated forms of data to filter the training samples from the uncertain regions. We adopt non-maximum suppression (NMS) with IoU threshold of 0.6 for post-processing in all experiments.
\section{Experiments and Results}

\subsection{Dataset and Evaluation Metrics}
\textbf{Dataset:}. We collect a total of 2239 CT images from patients with rib fractures, of which 685 were box-annotated data, 450 were dot-annotated data, and 1104 were unlabeled data. Then we divide the 685 cases of box annotation data into training dataset (224), validation dataset (151), and testing dataset (310).
\textbf{Evaluation Metrics:} The mean Average Precision (mAP) from AP40 to AP75 with an interval of 5 \footnote{\url{https://www.kaggle.com/c/rsna-pneumonia-detection-challenge}} and AP50 \footnote{\url{https://cocodataset.org/\#detection-eval}} are adopted as the evaluation metrics.

\subsection{Comparison with State-of-the-art Methods}
To the best of our knowledge, few studies have been proposed to simultaneously leverage the box-labeled data, dot-labeled data, and unlabeled data for object detection. Hence, we first train the supervised object detection model FCOS \cite{tian2020fcos} as the baseline model. Then we implement several semi-supervised methods including STAC \cite{sohn2020simple}, Unbiased Teacher \cite{liu2021unbiased}, $\Pi$ Model\cite{laine2016temporal} and AALS \cite{wang2021knowledge}. Note that, we enable the learning from the dot labeled data by training the supervised or semi-supervised model with only the annotated positive points. 

\begin{table}[]
\centering
\caption{Quantitative comparisons with different methods on the testing dataset.}
\begin{tabular}{c|ccc|ccc}
\hline
\multirow{2}{*}{Method} & \multicolumn{3}{c|}{\#Number of CT Scans} & \multicolumn{2}{c}{Metrics} \\ \cline{2-6} 
                        & \multicolumn{1}{c|}{\begin{tabular}[c]{@{}c@{}}Box-labeled\end{tabular}} & \multicolumn{1}{c|}{\begin{tabular}[c]{@{}c@{}}Dot-labeled\end{tabular}} & \begin{tabular}[c]{@{}c@{}}Unlabeled \end{tabular} & mAP     & AP50    \\ \hline
FCOS \cite{tian2020fcos}                   & \multicolumn{1}{c|}{224}                                                         & \multicolumn{1}{c|}{0}                                                           & 0                                                         & 39.9    & 53.7    \\ \hline

FCOS \cite{tian2020fcos}                 & \multicolumn{1}{c|}{224}      & \multicolumn{1}{c|}{450}                                                         & 0                                                         & 41.3   & 54.4   \\

ORF-Net                     & \multicolumn{1}{c|}{224}                                                         & \multicolumn{1}{c|}{450}                                                         & 0                                                         & \textbf{42.3}    & \textbf{56.3}    \\
 \hline
STAC \cite{sohn2020simple}                    & \multicolumn{1}{c|}{224}                                                         & \multicolumn{1}{c|}{450}                                                         & 1104                                                         & 40.0    & 56.1  \\

UT \cite{liu2021unbiased}                    & \multicolumn{1}{c|}{224}                                                         & \multicolumn{1}{c|}{450}                                                         & 1104                                                         & 42.6    & 56.3  \\
$\Pi$ Model \cite{laine2016temporal}     & \multicolumn{1}{c|}{224}                                                         & \multicolumn{1}{c|}{450}    & 1104  & 42.9    & 56.3  \\
AALS \cite{wang2021knowledge}  & \multicolumn{1}{c|}{224}                                                         & \multicolumn{1}{c|}{450}                                                         & 1104                                                         & 43.4    & 57.2  \\
ORF-Net                     & \multicolumn{1}{c|}{224}                                                         & \multicolumn{1}{c|}{450}                                                         & 1104                                                      & \textbf{44.3}    & \textbf{59.1}  \\ \hline
\end{tabular}
\label{tab:stateoftheart}
\end{table}

\begin{table}[]
\centering
\caption{Ablation studies on the validation dataset.}
\begin{tabular}{c|ccc|ccc}
\hline
\multirow{2}{*}{Method} & \multicolumn{3}{c|}{\#Number of CT Scans} & \multicolumn{2}{c}{Metrics} \\ \cline{2-6} 
                        & \multicolumn{1}{c|}{\begin{tabular}[c]{@{}c@{}}Box-labeled\end{tabular}} & \multicolumn{1}{c|}{\begin{tabular}[c]{@{}c@{}}Dot-labeled\end{tabular}} & \begin{tabular}[c]{@{}c@{}}Unlabeled\end{tabular} & mAP     & AP50    \\ \hline
FCOS \cite{tian2020fcos}                   & \multicolumn{1}{c|}{224}                                                         & \multicolumn{1}{c|}{0}                                                           & 0                                                         & 36.3    & 51.8     \\
\hline

ORF-Net (SGM)                & \multicolumn{1}{c|}{224}      & \multicolumn{1}{c|}{450}                                                         & 0                                                         & 38.8   & 53.7    \\
ORF-Net (IGM)                  & \multicolumn{1}{c|}{224}      & \multicolumn{1}{c|}{450}                                                         & 0                                                         & \textbf{39.9}   & 55.2    \\
ORF-Net (IGM+CA)                 & \multicolumn{1}{c|}{224}      & \multicolumn{1}{c|}{450}                                                         & 0                                                         & \textbf{39.9}   & \textbf{56.6}    \\
 \hline
ORF-Net (SGM)                     & \multicolumn{1}{c|}{224}                                                         & \multicolumn{1}{c|}{450}                                                         & 1104                                                      & 40.2    & 56.6       \\
ORF-Net (IGM)                    & \multicolumn{1}{c|}{224}                                                         & \multicolumn{1}{c|}{450}                                                         & 1104                                                      & 41.6    & 57.8        \\
ORF-Net (IGM+CA)                    & \multicolumn{1}{c|}{224}                                                         & \multicolumn{1}{c|}{450}                                                         & 1104                                                      & \textbf{42.8}    &\textbf{57.9}      \\ \hline
\end{tabular}
\label{tab:ablationstudy}
\end{table}
\begin{figure}[t]
\centering
	\includegraphics[width=\textwidth]{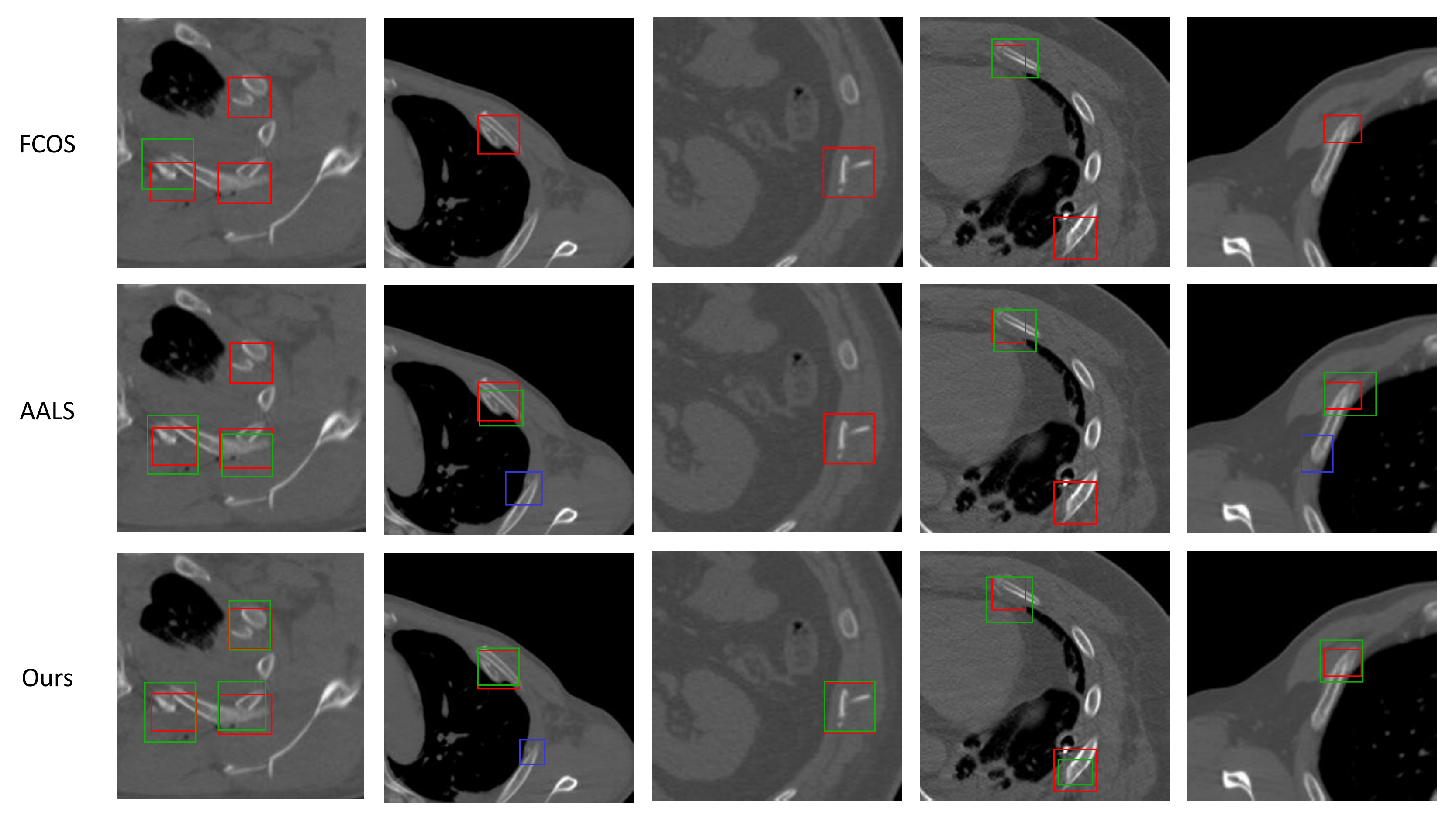}
	\caption{ Qualitative comparisons of the FCOS \cite{tian2020fcos}, AALS \cite{wang2021knowledge}, and Our proposed method on the testing set. Red rectangles stand for ground truths, green rectangles stand for true positives, and blue rectangles stand for false positives.}
		\label{visual}
\end{figure}

\noindent\textbf{Quantitative Results:} As shown in Table~\ref{tab:stateoftheart}, by simply training the annotated positive samples from the dot labeled data, the fully supervised method FCOS \cite{tian2020fcos} can also achieve an improvement of $1.4\%$ and $0.7\%$ on mAP and AP50. Our method achieved a large improvement on both mAP and AP50 ($2.4\%$, $2.6\%$), which demonstrated the effectiveness of the proposed ORF-Net on leveraging the dot-labeled data. 
By training with all the different forms of annotation, all the semi-supervised methods improved the fully supervised baseline, demonstrating effectiveness in utilizing the unlabeled data.
The proposed method outperformed all other methods with at least $0.9\%$ in mAP, and $1.9\%$ in AP50, demonstrating the efficacy of omni-supervised learning on the task of rib fracture detection.

\noindent\textbf{Ablation Study:}
For a better comparison, we also implement a self-guided map (SGM) based label assignment strategy which uses its own classification scores and annotations to conduct the label assignment.
As shown in Table~\ref{tab:ablationstudy}, compared with the SGM based label assignment strategy, the IGM based label assignment strategy has an improvement of $1.\%$ and $1.5\%$ on mAP and AP50 with the dot labeled data, and an improvement of $1.4\%$ and $1.5\%$ on the mAP and AP50 with the dot labeled data and unlabeled data, demonstrating the effectiveness of the proposed IGM-based label assignment strategy.  
For the confidence-aware (CW) loss, we observe an improvement of $1.4\%$ on AP50 with the dot labeled data, and an improvement of $1.2\%$ on the mAP with both the dot labeled data and unlabeled data, which validate the efficiency of the proposed soft regularization.

\noindent\textbf{Qualitative Results:}
We also visualize the predictions generated by FCOS, AALS, and ORF-Net in Fig. \ref{visual}. As illustrated, our model predicts more accurate rib fractures than FCOS and AALS, demonstrating the efficacy of the proposed method on leveraging different annotated forms of data. 

\section{Conclusion}
In this paper, we present an omni-supervised learning method for rib fracture detection from chest CT scans. The proposed omni-supervised network could dynamically conduct the label assignment for the different annotated forms of data and thus exploit the supervision of various levels to further improve the detection performance. 
Extensive experiments on the testing dataset demonstrate the efficiency of our method in utilizing the various granularities of annotations.
Moreover, the proposed method is general and can be easily extended to other tasks of object detection.

\subsubsection{Acknowledgement.} 
This work was supported by Key-Area Research and Development Program of Guangdong Province, China under Grant 2020B010165004, Hong Kong RGC TRS Project No. T42-409/18-R, and Shenzhen Science and Technology Innovation Committee Funding (Project No. SGDX20210823103201011).

%
%
\bibliographystyle{paper654}
\bibliography{paper654}

\end{document}